\pdfoutput=1

\documentclass[11pt]{article}
\usepackage{EACL2023}

\usepackage{times}
\usepackage{latexsym}

\usepackage{multirow}

\usepackage{subcaption}

\usepackage[T1]{fontenc}

\usepackage[utf8]{inputenc}
\usepackage{csquotes}

\usepackage{microtype}

\usepackage{inconsolata}

\newcommand\mscriptsize[1]{\mbox{\scriptsize\ensuremath{#1}}}

\title{Emotion Recognition based on Psychological Components in Guided Narratives for Emotion Regulation}

 \author{Gustave Cortal$^{1}$,
        Alain Finkel$^{1,4}$,
        Patrick Paroubek$^{2}$,
         Lina Ye$^{3}$ \\
$^1$Univ. Paris-Saclay, CNRS, ENS Paris-Saclay, LMF, 91190, Gif-sur-Yvette, France\\
    $^2$Univ. Paris-Saclay, CNRS, LISN, 91400, Orsay, France\\
    $^3$Univ. Paris-Saclay, CNRS, ENS Paris-Saclay, CentraleSupélec, 91190, Gif-sur-Yvette, France\\
    $^4$Institut Universitaire de France, France\\
          \texttt{\{gustave.cortal, alain.finkel\}@ens-paris-saclay.fr}, \\ \texttt{pap@limsi.fr}, \texttt{lina.ye@centralesupelec.fr}}

\begin{document}
\maketitle
\begin{abstract}

Emotion regulation is a crucial element in dealing with emotional events and has positive effects on mental health. This paper aims to provide a more comprehensive understanding of emotional events by introducing a new French corpus of emotional narratives collected using a questionnaire for emotion regulation. We follow the theoretical framework of the Component Process Model which considers emotions as dynamic processes composed of four interrelated components (\textsc{behavior}, \textsc{feeling}, \textsc{thinking} and \textsc{territory}). Each narrative is related to a discrete emotion and is structured based on all emotion components by the writers. We study the interaction of components and their impact on emotion classification with machine learning methods and pre-trained language models. Our results show that each component improves prediction performance, and that the best results are achieved by jointly considering all components. Our results also show the effectiveness of pre-trained language models in predicting discrete emotion from certain components, which reveal differences in how emotion components are expressed. 

\end{abstract}

\section{Introduction}

\begin{table*}[!htb]
    \centering
    
\begin{tabular}{lp{0.78\textwidth}}
\hline
 
                   Component &
                 Answer \\
 
\hline
          \textsc{behavior} & I'm giving a lecture on a Friday morning at 8:30. A student goes out and comes back a few moments later with a coffee in his hand. \\
\textsc{feeling} & My heart is beating fast, and I freeze, waiting to know how to act. \\
  \textsc{thinking} & I think this student is disrupting my class. \\
\textsc{territory} & The student attacks my ability to be respected in class. \\
 
\hline
\end{tabular}
\captionof{table}{Example of an emotional narrative structured according to emotion components. The writer identified that he was angry.}
\label{tab:description_corpus}
\end{table*}

Emotion analysis in text consists of associating an emotion from a predefined set (e.g. \textit{fear}, \textit{joy}, \textit{sadness}) to a textual unit (e.g. word, clause, sentence). Several psychological theories are used to define the emotion classes to be predicted. Basic emotion theories \citep{ekman} consider discrete emotions shared by all, as they may have innate neural substrates and universal behavioral phenotypes.  Dimensional theories \citep{russel}  define emotions through affective dimensions, such as the degree of agreeableness (\textit{valence}) and the degree of physiological activation (\textit{arousal}). 

Previous studies \citep{bostan-klinger-2018-analysis} have conducted analyses on various corpora for emotion classification in text. Most of them neglect the existing psychological knowledge about emotions, which can be used to clarify what an emotion is and how it can be caused. To the best of our knowledge, only a few approaches incorporate cognitive psychological theories to classify emotions in texts. These include a knowledge-base-oriented modeling of emotional events \citep{cambria}, a corpus annotated according to dimensions of cognitive appraisal of events \citep{10.1162/coli_a_00461}, an annotation scheme for emotions inspired by psycholinguistics \citep{etienne-etal-2022-psycho}, and the identification of emotion component classes \citep{casel-etal-2021-emotion} according to the Component Process Model (CPM) \citep{component} in cognitive psychology.

These papers, like ours, are based on the cognitive appraisal theory \citep{lazarus}, which posits that emotions arise from the evaluation of an event based on various cognitive criteria, such as \textit{relevance}, \textit{implication}, \textit{coping}, and \textit{normative significance}. The CPM is rooted in this theory and defines emotion as a set of cognitive appraisals that modulate the expression of five components in reaction to an event (\textit{cognitive appraisal}, \textit{physiological response}, \textit{motor expression}, \textit{action tendency}, and \textit{subjective feeling}). Our chosen components are closely related to the components originally proposed in the CPM. In this paper, we follow the theoretical framework of the CPM by considering emotions as dynamic processes composed of four interrelated components: \textsc{behavior} (\enquote{I'm giving a lecture}), \textsc{feeling} (\enquote{My heart is beating fast}), \textsc{thinking} (\enquote{I think he's disrupting my lecture}) and \textsc{territory} (\enquote{He attacks my ability to be respected}) proposed by \citet{Finkel2022}. In our corpus, each narrative is structured by the writers according to these components. Table \ref{tab:description_corpus} shows an example of a structured narrative.

We rely on the same assumptions made by \citet{casel-etal-2021-emotion}, namely that emotions in a text are expressed in several ways. Emotion components are associated with different linguistic realizations. In this paper, we study how emotions are expressed through components by introducing a new French corpus composed of emotional narratives. Narratives were collected with a questionnaire following a new  psychological method, called Cognitive Analysis of Emotions \citep{Finkel2022}, which aims to modify (negative) representations of an emotional event to help people better regulate their emotions.  Our corpus is structured according to emotion components and contains 812 narratives, corresponding to 3082 answers. Each narrative contains several answers, and each answer corresponds to a single component.

In this paper, we describe the annotation of our corpus and evaluate traditional machine learning methods and pre-trained language models for discrete emotion classification based on emotion components. To the best of our  knowledge, this work is the first to study the interaction between linguistic realizations of components for emotion classification. We aim to answer several questions: does a component influence emotion prediction and, if so, does it increase or decrease performance? Does each component contribute equally or unequally to the prediction? Does considering all components lead to the best performance?

\paragraph{Contributions}We present a new French corpus composed of emotional narratives structured according to four components (\textsc{behavior}, \textsc{feeling}, \textsc{thinking} and \textsc{territory}). Each narrative is related to a discrete emotion and is structured based on all emotion components by the writers, allowing us to study the interaction of components and their impact on emotion classification. We evaluate the influence of components on emotion classification using traditional machine learning methods and pre-trained language models (CamemBERT). Our results show that each component improves prediction performance, and that the best results are achieved by jointly considering all components. Our results also show that CamemBERT effectively predict discrete emotion from \textsc{thinking}, but do not improve performance from \textsc{feeling} compared to traditional machine learning approaches, which reveal differences in how emotion components are expressed. We believe that our analysis can provide a further insight into the semantic core of emotion expressions in text.

\section{Background and Related Work}

\subsection{Psychological Theories of Emotion}

\paragraph{Discrete and Continuous theories}Among emotion theories, we can distinguish between those that suppose the existence of a finite number of distinct basic emotions and those considering that emotion has several dimensions.
The basic emotion theories list several emotions common to human beings, such as Ekman's universal emotions (\textit{sadness}, \textit{joy}, \textit{anger}, \textit{fear}, \textit{disgust}, and \textit{surprise}) \citep{ekman} and Plutchik's wheel of emotions \citep{plutchik}. Instead of categorizing an emotion according to a discrete set, dimensional theories consider emotion as a point in a multidimensional Euclidean space. For example, \citet{russel} consider emotions along three dimensions: an emotion is identifiable according to its degree of agreeableness (\textit{valence}), its degree of physiological activation (\textit{arousal}), and its degree of felt control (\textit{dominance}).

 \paragraph{Appraisal theories}The cognitive appraisal theory \citep{lazarus} identifies cognitive dimensions of emotion, considered criteria for evaluating an event. For example, it considers that an individual evaluates how an event helps him or her in satisfying a need or accomplishing a goal. There are other appraisal criteria, such as the ability to cope with an event based on resources available to the individual. The type and intensity of an emotion provoked by an event depend on the result of cognitive appraisals.

\paragraph{Component Process Model}Cognitive appraisals are integrated in the Component Process Model (CPM) \citep{component}. It considers emotion as the expression of several components (\textit{cognitive appraisal}, \textit{physiological response}, \textit{motor expression}, \textit{action tendency}, and \textit{subjective feeling}) that synchronize in reaction to an event. The cognitive appraisals of an event modulate the expression of components. For example, during an exam, I evaluate my ability to solve an exercise; I think I do not have the skills to solve it and will get a bad mark (\textit{cognitive appraisal}). I panic (\textit{subjective feeling}), I sweat (\textit{physiological response}), my legs shake (\textit{motor expression}), I feel like getting up and running away from the classroom (\textit{action tendency}). In this text, we can infer that I am afraid (\textit{fear}). Our corpus explores the interaction between linguistic realizations of components. Despite being closely related, our components proposed by the Cognitive Analysis of Emotion differ from the original ones presented by the CPM.

\paragraph{Cognitive Analysis of Emotion}The Cognitive Analysis of Emotion \citep{Finkel2022} is a cognitive appraisal theory that explores the basic emotions (\textit{anger}, \textit{fear}, \textit{joy}, and \textit{sadness}) with their corresponding behavioral (\textsc{behavior}), physiological (\textsc{feeling}), and cognitive (\textsc{thinking} and \textsc{territory)} components. Like other psychological and neuroscientific theories, it assumes that the mind processes emotional information, in order to prepare for and take appropriate action. If the information is not processed satisfactorily according to an individual's values, beliefs, or goals, the mind may repress, block, or loop, leading to unsatisfactory outcomes. The Cognitive Analysis of Emotion uses the CPM to reorganize the narrative of experienced emotional events. This process helps individuals better understand and regulate their emotions, as well as prepare for necessary actions. It provides a method for understanding emotions that can modify negative representations of emotional events. The narratives are categorized using a questionnaire, presented in Section \ref{corpus_annotation}. \citet{cortal:hal-03805702} introduce the use of natural language processing to automate parts of the Cognitive Analysis of Emotion.

\subsection{Emotion Analysis in Text}

Most methods for analyzing emotions in text focus on either the classification of discrete emotional states \citep{bostan-klinger-2018-analysis} or the recognition of affective dimensions such as \textit{valence}, \textit{arousal}, and \textit{dominance} \citep{buechel-hahn-2017-emobank}. 

\paragraph{Emotion Cause Extraction} Recently, some new studies aim to not only recognize the emotional state present in the text, but also the span of text that serves as its underlying cause. \citet{lee-etal-2010-text} introduce the Emotion Cause Extraction task and define it as the identification of word-level factors responsible for the elicitation of emotions within text. \citet{chen-etal-2010-emotion} analyze the corpus presented by \citet{lee-etal-2010-text} and suggest that clause-level detection may be a more suitable unit for detecting causes. \citet{xia-ding-2019-emotion} propose the Emotion-Cause Pair Extraction task, i.e., the simultaneous extraction of both emotions and their corresponding causes. Several extensional approaches have been proposed to address this task with better performance (\citet{ding-etal-2020-ecpe}, \citet{wei-etal-2020-effective}, \citet{ding-etal-2020-end}, \citet{chen-etal-2020-unified}, \citet{DBLP:conf/wassa/SinghHWM21}).

\paragraph{Structured Emotion Analysis} 

The goal of semantic role labelling \citep{gildea-jurafsky-2000-automatic} is to determine the participants involved in an action or event indicated by a predicate in a given sentence. For emotion analysis, the task shifts its focus from actions to emotional cues, which are words or expressions that trigger emotions. Emotion semantic role labelling consists of answering the question: \enquote{Who feels What, towards Whom, and Why?} \citep{campagnano-etal-2022-srl4e}. \citet{salif} annotate tweets during the 2012 U.S. presidential elections, \citet{bostan-etal-2020-goodnewseveryone} annotate news headlines and \citet{reman} annotate literary paragraphs. They identify emotion cues with the corresponding emotion experiencers, causes and targets. \citet{campagnano-etal-2022-srl4e} propose a unified annotation scheme for different emotion-related semantic role corpora, including those presented previously. To the best of our knowledge, the only French language studies that address the identification of emotion-related semantic roles are the corpus for recognizing emotions in children's books \citep{etienne-etal-2022-psycho}, the corpus for extremist texts \citep{dragos-etal-2022-angry}, and the Défi Fouille de Textes campaign \citep{paroubek-etal-2018-deft2018}, which annotates tweets related to transportation in the Île-de-France region.

\paragraph{Appraisal Theories for Emotion Analysis} 

A few approaches incorporate cognitive psychological theories to classify emotions in text. The ISEAR project \citep{Scherer1994} compiles a textual corpus of event descriptions. However, they focus on the existence of emotion components, but not on the linguistic expression of emotion components. \citet{cambria} identify event properties including people’s goals for sentiment analysis using a knowledge-base-oriented approach. \citet{10.1162/coli_a_00461} compile a corpus that considers the cognitive appraisal of events from both the writer and reader perspectives. Very few studies focus on emotion component analysis. \citet{kim-klinger-2019-analysis} analyze the communication of emotions in fan fiction through some variables related to emotion components such as facial and body posture descriptions, subjective sensations, and spatial relations of characters. \citet{casel-etal-2021-emotion} annotate existing literature and Twitter emotion corpora with emotion component classes based on the CPM. However, not all emotion components are expressed to characterize an emotional event. In our corpus, each narrative is structured based on all emotion components, allowing us to study the interaction of components and their impact on emotion classification.

\citet{Mentrey2022} represents the pioneering effort in examining the interaction of components for discrete emotion prediction. However, their annotation approach deviates from ours. They use a scale ranging from 1 to 7 to solicit annotators' agreement with predefined descriptions (e.g. \enquote{To what extent did you feel calm?}). This approach disregards the linguistic manifestation of emotion components. In contrast, our questionnaire employs open-ended questions to gather the linguistic expression of emotional events, enabling the application of natural language processing techniques. 

\section{Corpus Creation}

\begin{table*}[!htb]
    \begin{subtable}{.5\linewidth}
      \centering
        
\begin{tabular}{lcc|lc}
\hline
 
                   Component &
                 $\#A$ & $\overline{t_A}$ & Emotion & \% \\
 
\hline
          \textsc{behavior} & 802 & 82 & \textit{Anger} & 52 \\
\textsc{feeling} & 799 & 27 & \textit{Fear} & 36 \\
  \textsc{thinking} & 799 & 54 & \textit{Sadness} & 14 \\
\textsc{territory} & 682 & 34 & \textit{Joy} & 11 \\
\hline
\end{tabular}
        \caption{Entire corpus (\underline{Total}).}
    \end{subtable}
    \begin{subtable}{.5\linewidth}
      \centering
        
\begin{tabular}{lcc|lc}
\hline
 
                   Component &
                 $\#A$ & $\overline{t_A}$ & Emotion & \% \\
 
\hline
          \textsc{behavior} & 392 & 93 & \textit{Anger} & 48 \\
\textsc{feeling} & 392 & 26 & \textit{Fear} & 32 \\
  \textsc{thinking} & 392 & 59 & \textit{Sadness} & 10 \\
\textsc{territory} & 392 & 38 & \textit{Joy} & 10 \\
\hline
\end{tabular}
        \caption{Subset of \underline{Total} for the emotion classification task (\underline{Emotion}).}
    \end{subtable} 
    \caption{Number of answers ($\#A$), average number of tokens for answers ($\overline{t_A}$) and distribution of emotion classes. For \underline{Total}, a questionnaire can correspond to more than one emotion class.}
    \label{tab:corpus}
\end{table*}

\begin{table}[!htb]
    \centering
\begin{tabular}{lccccc}
\hline
 & $\#N$ & $\overline{t_N}$ & $\#A$ & \% completion \\
\hline
\underline{Total} & 812 & 190 & 3082 & 61 \\
\underline{Emotion} & 392 & 216 & 1568 & 100 \\ 
\hline
\end{tabular}
   \caption{Number of narratives ($\#N$), average number of tokens for narratives ($\overline{t_N}$), number of answers ($\#A$) and completion rate for questionnaires. Statistics for the entire corpus (\underline{Total}) and the subset for the emotion classification task (\underline{Emotion}).}
   \label{tab:general_corpus}
\end{table}

\subsection{Corpus Annotation\label{corpus_annotation}}

In a Cognitive Analysis session, the participants, who wish to manage their emotions better, write a narrative of an experienced emotional event with identified characters in a given place and time. The writer first identifies the basic emotion he/she has experienced, then he/she structures the narratives according to emotion components by filling in a questionnaire. The writer also describes the actions that could have been performed but that he/she had not considered or that he/she had forbidden himself/herself to do during the emotional event. We do not consider this last action part of the questionnaire in our study, as we are only interested in the emotion components.\footnote{We point out that, in this paper, we only study the linguistic realizations of emotion components.} Table \ref{tab:description_corpus} shows a structured narrative based on  components described by \citet{Finkel2022}. We provide a summary below :

\begin{itemize}

\item \textsc{Behavior}: the writer describes the observable behaviors of himself/herself and others. They are identified by answering \enquote{Who did what?} and \enquote{Who said what?}. The writer also provides the context of an emotional event, such as location and date. 
\item \textsc{Feeling}: the writer expresses his/her physical feelings during the emotional event.

\item \textsc{Thinking}: the writer reports what he/she thought during the emotional event.
\item \textsc{Territory}: the writer describes whether his/her needs are satisfied or not by analyzing the different cognitive appraisals that he/she thinks he/she has made during the emotional event. The Cognitive Analysis of Emotion considers that an emotion arises when we evaluate an event that invalidates or confirms our model of the world, the latter containing territories associated with our needs. Territories are concrete objects such as an individual body or home, or abstract objects such as individual values, beliefs, or self-image.

\end{itemize}

Using the questionnaire, a writer categorizes an emotional narrative by considering four emotion components (\textsc{behavior}, \textsc{feeling}, \textsc{thinking}, and \textsc{territory}) proposed by \citet{Finkel2022}, closely related to components originally proposed by the CPM. For example, \textsc{feeling} may contain \textit{physiological responses} (\enquote{My heart is beating fast}) and \textit{motor expressions} (\enquote{I feel I am smiling}). \textsc{Thinking} may contain \textit{action tendencies} (\enquote{I felt like hitting him}) and \textit{subjective feelings} (\enquote{I was relaxed}). \textsc{territory} provides information on criteria involved in the \textit{cognitive appraisal} of an event (\enquote{The student attacks my ability to be respected in class}).

We point out that compared to previous studies on emotion component analysis \citep{casel-etal-2021-emotion, Mentrey2022}, our corpus contains linguistic realizations of all components for each emotional narrative, providing a more comprehensive understanding of emotional events. \citet{Mentrey2022} do not consider linguistic realizations of components, and \citet{casel-etal-2021-emotion} do not consider all components for each emotional event, hence they cannot study the interaction of components. Moreover, our corpus is annotated by the writers of narratives themselves, rather than external annotators, as in \citet{casel-etal-2021-emotion}. An interesting direction for future research would involve the incorporation of external annotations into our corpus to conduct a comparative analysis between the writer's perspective and that of the reader.

\subsection{Corpus Statistics}

Practitioners trained in Cognitive Analysis of Emotion manually collected questionnaires from individuals who chose to participate in emotion regulation trainings between 2005 and 2022. During these years, the format of questionnaires has changed several times, as well as the instructions given. All questionnaires were converted into a standard format. Each questionnaire is completed by a single person and corresponds to a narrative related to a discrete emotion. We did not collect specific data on the writers. Most of them are master's students (20 to 22 years old), doctoral students (22 to 30 years old) and teachers (25 to 50 years old, with an average around 30) studying or working in France, and who have given their consent for the questionnaires to be collected and processed. 

Narratives are disidentified using a named entity recognition model.\footnote{\url{https://huggingface.co/Jean-Baptiste/camembert-ner}} We then manually verify and correct the automatic disidentification. Specific tokens replace personal names, organizations, dates, and locations to preserve the privacy of writers. We delete empty answers containing less than 3 tokens. 

Our corpus is composed of 812 unique questionnaires, for a total of 3082 answers (\underline{Total}). Each answer is related to a single component. We introduce a subset (\underline{Emotion}) of our entire corpus (\underline{Total}) composed of questionnaires with all components filled in and corresponding to a single emotion class. For the emotion classification task, described in the next section, we use the \underline{Emotion} subset.

Corpus statistics obtained with SpaCy \citep{spacy2} are illustrated for each component in Table \ref{tab:corpus}. Although a questionnaire corresponds to one primary emotion class, sometimes writers indicate experiencing other secondary emotion classes. Table \ref{tab:corpus} also shows the distribution of emotion classes. The dominance of negative emotions is expected; writers usually fill in a questionnaire when they want to better deal with a distressing event. Table \ref{tab:general_corpus} shows general statistics for \underline{Total} and \underline{Emotion}.

\section{Experiments and Results}

\begin{table*}[!htb]
    \centering
\begin{tabular}{llllllll}
\hline
&\multicolumn{3}{c}{Logistic Regression}&\multicolumn{3}{c}{CamemBERT} \\

                   Component &  Precision &     Recall &   $F_1$ &  Precision &     Recall &   $F_1$ \\
\hline
              All  & 71.2\,\mscriptsize{(2.6)} & 69.1\,\mscriptsize{(2.2)} & 67.8\,\mscriptsize{(2.3)} & \textbf{85.1} & \textbf{84.8} & \textbf{84.7} \\
              Without \textsc{behavior}   & 77.4\,\mscriptsize{(2.3)} & 75.8\,\mscriptsize{(2.4)} & 74.5\,\mscriptsize{(2.6)} & 80.3 & 79.8 & 79.7 \\
              Without \textsc{feeling}  & 64.3\,\mscriptsize{(1.9)} & 61.5\,\mscriptsize{(1.2)} & 61.3\,\mscriptsize{(2.2)} & 81.6 & 79.8 & 79.9  \\
              Without \textsc{thinking}  & 70.9\,\mscriptsize{(1.8)} & 69.1\,\mscriptsize{(2.0)} & 68.3\,\mscriptsize{(2.2)} & 79.6 & 78.5 & 78.7 \\
              Without \textsc{territory}  & 64.3\,\mscriptsize{(4.1)} & 64.5\,\mscriptsize{(2.4)} & 62.3\,\mscriptsize{(2.8)} & 78.7 & 78.5 & 78.6 \\
          Only \textsc{behavior}  & 52.1\,\mscriptsize{(3.5)} & 54.6\,\mscriptsize{(2.9)} & 51.7\,\mscriptsize{(2.9)} & 68.4  & 67.1 & 66.6 \\
Only \textsc{feeling}  & 69.6\,\mscriptsize{(1.5)} & 68.9\,\mscriptsize{(2.1)} & 68.4\,\mscriptsize{(2.0)} & 67.8 & 68.4 & 67.7 \\
  Only \textsc{thinking}  & 50.1\,\mscriptsize{(3.4)} & 53.8\,\mscriptsize{(2.3)} & 50.6\,\mscriptsize{(2.7)} & 70.5 & 70.1 & 70.1 \\
              Only \textsc{territory}  & 68.2\,\mscriptsize{(1.8)} & 66.8\,\mscriptsize{(2.2)} & 66.6\,\mscriptsize{(2.3)} & 71.4 & 68.4 & 68.9 \\
\hline
\end{tabular}
    \caption{Scores (± std) for discrete emotion classification based on components.}
    \label{tab:pred_emotion}
\end{table*}

\subsection{Methods}

In this study, we aim to examine the interaction between linguistic realizations of emotion components through traditional machine learning methods and pre-trained language models. Our corpus is unique in that it provides multiple components for each emotional event, enabling us to investigate the interaction of components and their impact on emotion classification. Our research questions include: does the presence of a component impact emotion prediction? Does considering all components result in the best prediction performance? We answer the same questions posed by \citet{Mentrey2022}, but we focus on the linguistic expression of emotional events, instead of the existence of described event properties.

\paragraph{Traditional machine learning methods}We train logistic regressions, support vector machines, and random forests on our corpus represented as a bag-of-words (unigrams), averaged using the TF-IDF method. The words are pre-processed through lemmatization using SpaCy. To prevent bias, we remove terms directly related to the emotion classes (e.g. \enquote{fear}, \enquote{anger}, \enquote{sad}, \enquote{joy}). For model evaluation, we perform a five-fold cross-validation, and we calculate $F_1$ score, recall, and precision using a weighted mean.\footnote{As the emotion class distribution is imbalanced.} For training our models, we use Scikit-learn \citep{scikit-learn} with default hyperparameters. 

\paragraph{Pre-trained language models}We fine-tune a transformers-based model \citep{vaswani2017attention} using the distilled version \citep{delestre:hal-03674695} of CamemBERT \citep{martin-etal-2020-camembert}, a BERT model \citep{devlin-etal-2019-bert} for the French language. We use the raw answers, but we also remove terms directly related to the emotion classes to prevent bias. The corpus is split into 80\% for training and 20\% for evaluation. We train models for 5 epochs using PyTorch \citep{NEURIPS2019_9015} and HuggingFace's Transformers \citep{wolf-etal-2020-transformers}, with model parameters for each epoch saved. We select the model with the highest $F_1$ score on the evaluation data. Training hyperparameters and fine-tuned CamemBERT weights are publicly available on HuggingFace.\footnote{\url{https://huggingface.co/gustavecortal/distilcamembert-cae-all}}

\subsection{Emotion Classification}

In this study, we aim to investigate the impact of component interaction on discrete emotion classification. We train models on all components at once, on all but one component at once, and on a single component. To account for multiple components, we concatenate their respective answers. For example, \enquote{Only \textsc{territory}} models are trained on \textsc{territory}, \enquote{Without \textsc{Behavior}} models are trained on all components except \textsc{behavior} and \enquote{All} models are trained on all components, which represent an entire narrative. Models are trained on the \underline{Emotion} subset.

\paragraph{Results}Results are shown in Table \ref{tab:pred_emotion}. We do not show the performance of support vector machines and random forests since they perform worse than logistic regressions. The best results are achieved when all components are considered simultaneously, as indicated by the highest $F_1$ (84.7) with CamemBERT \enquote{All}. The results of CamemBERT models with the removal of individual components show a decrease in performance compared to CamemBERT \enquote{All}, with a decrease in $F_1$ ranging from -4.8 for \enquote{Without \textsc{feeling}} to -6.1 for \enquote{Without \textsc{territory}}. Hence, each component is relevant for classifying discrete emotions. Our findings lend support to Scherer's hypothesis \citep{component} that an emotional event is characterized by the synchronization of emotion components. This result is not self-evident, as individual components may convey conflicting information regarding the emotion classification task. Our results, coming from a natural language processing perspective, are consistent with those of \citet{Mentrey2022}, who studied the interaction of components for discrete emotion prediction from the existence of described event properties. 

In general, CamemBERT models show improved performance relative to logistic regressions, which is in line with expectations. However, the improvement is inconsistent across the models that only considered a single component, ranging from -0.7 for \enquote{Only \textsc{feeling}} to +19.5 for \enquote{Only \textsc{thinking}}. Our results show an important increase in $F_1$ for \enquote{Only \textsc{behavior}} (+14.9) and \enquote{Only \textsc{thinking}} (+19.5), whereas \enquote{Only \textsc{territory}} shows a slight increase (+2.3) and \enquote{Only \textsc{feeling}} shows a slight decrease (-0.7). We discuss these results which reveal ways in which components are expressed in a text. 

\paragraph{Discussions}

For emotions expressed through \textsc{territory} (+2.3 for \enquote{Only \textsc{territory}}), we believe that the way the question is asked to the writers influences strongly the way they answer, hence answers are biased due to the questionnaire format. For example, according to the Cognitive Analysis of Emotion, an attacked territory indicates that the corresponding emotion is \textit{anger} or \textit{fear}. Hence, the presence of only two unigrams, \enquote{territory} and \enquote{attack} can discriminate between \textit{anger} \textit{fear} and \textit{joy} \textit{sadness}, which can easily be performed by a logistic regression with TF-IDF features.

For emotions expressed through \textsc{behavior} (+14.9 for \enquote{Only \textsc{behavior}}), we believe that CamemBERT can discriminate the writer's behaviors from the behaviors of others characters in an emotional event, thus improving emotion prediction compared to logistic regressions.

CamemBERT improves performance for emotions expressed through \textsc{thinking} (+19.5 for \enquote{Only \textsc{thinking}}), while not having an important impact on performance for emotions expressed through \textsc{feeling} (-0.7 for \enquote{Only \textsc{feeling}}). Emotion expression modes \citep{Micheli2014}, studied in linguistics, could explain the differences in performance between logistic regressions and CamemBERT models trained on individual components. \citet{Micheli2014} presents a comprehensive study of French emotion denotation, examining the diverse mechanisms used to convey emotions in text. The study categorizes a vast array of heterogeneous markers into three emotion expression modes: emotions directly labeled by emotional words (\textit{labeled emotion}), emotions displayed through characteristics of utterances (\textit{displayed emotion}), and emotions illustrated by the description of a situation socially associated with an emotion (\textit{suggested emotion}).

We hypothesize that there is an important, yet unexplored, relationship between emotion expression modes and linguistic realizations of emotion components. For instance, \textsc{thinking} may include \textit{suggested emotions}, while \textsc{feeling} may include \textit{labeled emotions}. Classifying discrete emotions based on a \textit{suggested emotion} (e.g. \enquote{I think this student is disrupting my class}) would be more challenging compared to classifying discrete emotions from a \textit{labeled emotion} (e.g. \enquote{I am upset}). Understanding a \textit{suggested emotion} requires the understanding of the entire sentence and the sociocultural context of the emotional event, whereas understanding a \textit{labeled emotion} only requires identifying the relevant emotional words (\enquote{upset}), which can easily be performed by a logistic regression. Therefore, CamemBERT models are likely to outperform logistic regressions in terms of performance for emotions expressed through the \textit{suggested emotion} mode. This is due to CamemBERT's ability to encode the meaning of a sentence as a whole, as well as its pre-training that allows it to grasp the sociocultural context of an event, which logistic regression with TF-IDF features cannot do.

\subsection{Component Classification}

\begin{table}[!htb]
    \centering
\begin{tabular}{lllll}
\hline
Model &    Precision &      Recall &    $F_1$ \\
\hline
LR &  84.9\,\mscriptsize{(0.3)} &  84.3\,\mscriptsize{(0.3)} &  84.4\,\mscriptsize{(0.3)} \\
cBERT & \textbf{93.2} &  \textbf{93.0} &  \textbf{93.1} \\
\hline
\end{tabular}
    \caption{Scores ($\pm$ std) for emotion component classification. cBERT = CamemBERT.}
    \label{tab:app_auto}
\end{table}

We train traditional machine learning models and fine-tune CamemBERT to predict the emotion component class, i.e., whether an answer is a \textsc{behavior}, a \textsc{feeling}, a \textsc{thinking}, or a \textsc{territory}. Compared to the emotion classification task, models are trained on the entire corpus \underline{Total}.

Table \ref{tab:app_auto} show the results. We obtain great performances, logistic regression and CamemBERT can easily identify emotion component classes in our corpus. Training hyperparameters and fine-tuned CamemBERT weights are publicly available on HuggingFace.\footnote{\url{https://huggingface.co/gustavecortal/distilcamembert-cae-component}} We hope our corpus will benefit the research community for classifying components in text, a recent task introduced by \citet{casel-etal-2021-emotion}.

\section{Conclusion and Future Work}

Emotion regulation is a critical aspect of emotional events and has noteworthy implications for psychological well-being. In this paper, we aimed to provide a more comprehensive understanding of emotional events by introducing a French corpus of 812 emotional narratives (3082 answers). Our corpus was annotated following the Component Process Model and was collected using a recent psychological method for emotion regulation, named the Cognitive Analysis of Emotion. \citet{casel-etal-2021-emotion} were the first to annotate corpora with external annotators according to emotion components. Our corpus differs because each narrative is annotated by the writers and is structured according to all components (\textsc{behavior}, \textsc{feeling}, \textsc{thinking}, and \textsc{territory}), which allows for the study of their interaction.

We employed traditional machine learning methods and pre-trained language models (CamemBERT) to investigate the interaction of components for discrete emotion classification. Our results show that each component is useful for classifying discrete emotions, and that the model with the best performance considers all components, supporting Scherer's hypothesis \citep{component} that components synchronize during an emotional event.

Our results also show that CamemBERT effectively predict discrete emotion from \textsc{thinking}, but do not improve performance from \textsc{feeling} compared to traditional machine learning approaches, which reveal differences in how emotion components are expressed. We hypothesize that this may be explained by emotion expression modes studied in linguistics \citep{Micheli2014}. To test this hypothesis, we plan to annotate emotion expression modes in our corpus using a recent annotation scheme proposed by \citet{etienne-etal-2022-psycho}.

\section*{Limitations}

In our corpus, the distribution of emotion classes is imbalanced, which may bias the analyses, and notably impact the performance of trained models. Moreover, the data collected through a questionnaire may suffer from response bias, as the language used to describe an emotional narrative can be influenced by the questionnaire format and the elapsed time between the emotional event and its verbalization. We also point out that the linguistic expression of emotion does not necessarily capture the full extent of an emotional event, thus different from psychological or physiological studies on emotion~\cite{Simeng2019}. 

\section*{Ethics Statement}

In this paper, we collected data from individuals who attended emotion regulation trainings and provided consent for the collection and analysis of questionnaires. The corpus has not been published yet, as it is undergoing validation by the ethics committee of École Normale Supérieure Paris-Saclay.

By disidentifying our corpus, we have taken standard precautions to mitigate the introduction of biases into our models. Despite our efforts, it is possible that our models may still contain biases that we are not aware of. Our models are not intended for diagnostic purposes, and we do not provide automatic feedback to individuals for regulating their emotion, as we would need to be sure that such feedback does not have any adverse effects on individuals' mental health and, instead, facilitates improved emotion regulation.

\bibliography{anthology,custom}
\bibliographystyle{acl_natbib}

%\appendix

%\section{Example Appendix}
%\label{sec:appendix}

%This is a section in the appendix.

\end{document}